\pdfoutput=1

\documentclass[dvipsnames,11pt]{article}

\usepackage{acl}

\usepackage{times}
\usepackage{latexsym}
\usepackage[T1]{fontenc}
\usepackage[utf8]{inputenc}
\usepackage{microtype}
\usepackage{inconsolata}
\usepackage{graphicx}
\usepackage{amsmath}
\usepackage{bbm}
\usepackage{amssymb}
\usepackage{amsfonts}
\usepackage{booktabs}
\usepackage{lipsum}
\usepackage{multicol}
\usepackage{makecell}
\usepackage{rotating}
\usepackage{multirow}
\usepackage{utils}
\usepackage{caption}
\usepackage{subcaption}
\usepackage{xcolor}
\usepackage{enumitem}
\usepackage{linguex}
\usepackage{todonotes}
\usepackage[normalem]{ulem}
\usepackage{wasysym}
\usepackage{nicefrac}

\usepackage{comment}

\definecolor{customred}{HTML}{ac2c14}
\definecolor{customgreen}{HTML}{02ab8e}

\usepackage{csquotes}
\usepackage{changepage}
\renewenvironment{displayquote}
   {\begin{adjustwidth}{-1em}{-1em}\quote}
   {\endquote\smallskip\end{adjustwidth}}

\newcommand{\spose}{SPoSE}

\title{Characterizing the Role of Similarity in the\\Property Inferences of Language Models}

\author{
    \textbf{Juan Diego Rodriguez}$^\tau$ \quad \textbf{Aaron Mueller}$^{\nu, \iota}$ \quad \textbf{Kanishka Misra}$^{\bigstar, \chi}$\\
    $^\tau$The University of Texas at Austin \quad $^\nu$Northeastern University\\$^\iota$Technion -- Israel Institute of Technology\quad $^\chi$Toyota Technological Institute at Chicago\\
    \texttt{juand-r@utexas.edu}\quad \texttt{aa.mueller@northeastern.edu}\quad \texttt{kanishka@ttic.edu}
}

\begin{document}
\maketitle
\begin{abstract}

Property inheritance---a phenomenon where novel properties are projected from higher level categories (e.g., birds) to lower level ones (e.g., sparrows)---provides a unique window into how humans organize and deploy conceptual knowledge.~It is debated whether this ability arises due to explicitly stored taxonomic knowledge vs.\ simple computations of similarity between mental representations. %
How are these mechanistic hypotheses manifested in contemporary language models?
In this work, we investigate how LMs perform property inheritance with behavioral and causal representational analysis experiments.~We find that taxonomy and categorical similarities are not mutually exclusive in LMs' property inheritance behavior. That is, LMs are more likely to project novel properties from one category to the other when they are taxonomically related and at the same time, highly similar. 
Our findings provide insight into the conceptual structure of language models and may suggest new psycholinguistic experiments for human subjects.\footnote{Data and code available at \url{https://github.com/aaronmueller/lm-property-inheritance}.}
\end{abstract}

{\let\thefootnote\relax\footnotetext{\hspace{-0.2cm}$^{\bigstar}$Work partly done at UT-Austin before joining TTIC.}}

\section{Introduction}

Categories are fundamental to human semantic cognition.~Our knowledge of categories allows us to draw everyday inferences; a prominent example of which is \textit{property inheritance}, where properties are projected from a category to its members. For example, if we learn that \textit{dogs} have the T9 hormone, we can reasonably %
assume that \textit{corgis} also have the T9 hormone.~In cognitive psychology, an obvious and previously popular explanation for property inheritance is that it is the natural consequence of our minds organizing categories hierarchically into taxonomies \citep{collins1969retrieval, glass1974alternative, murphy2004big}.

At the same time, empirical evidence has called this assumption into question.~\citet{sloman1998categorical} showed that humans were highly sensitive to the similarity of the categories when performing property inheritance; for example, they were more likely to project novel properties from \textit{birds} to \textit{robins} (a more typical bird) than from \textit{birds} to \textit{penguins} (a less typical bird), despite them agreeing that both were members of the bird category. %
\citeauthor{sloman1998categorical} concluded that instead of explicitly using stored taxonomies, humans might simply be computing categorical similarities to demonstrate inheritance-compatible behavior. In such a case, a property is more likely to be shared between categories insofar as the similarity between them is high enough.

\begin{figure}[!t]
    \centering
    \includegraphics[trim=230 0 230 0, clip, width=\linewidth]{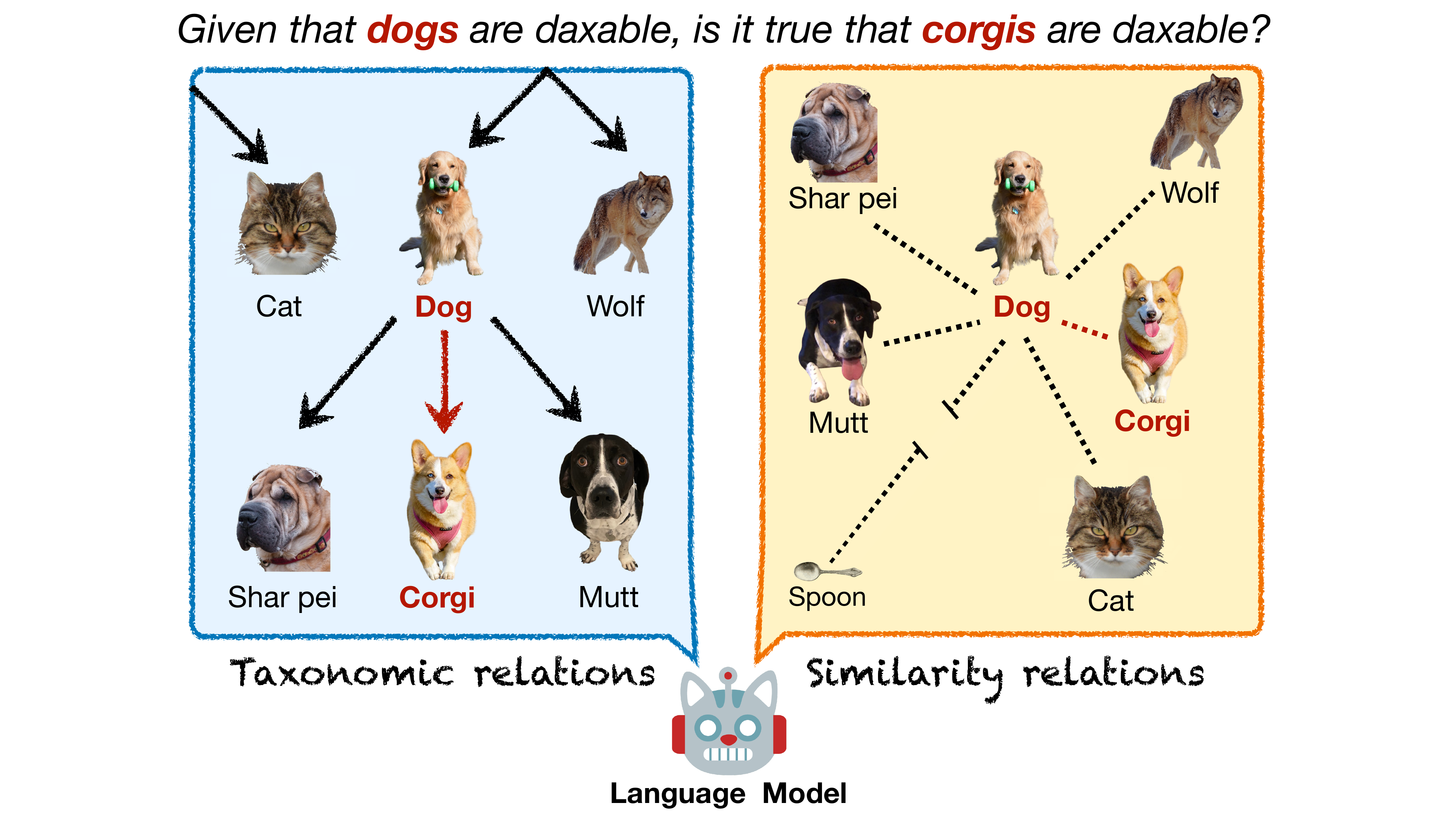}
    \caption{\textbf{Property inheritance} involves projecting properties from a category to its members. For example, if {\color{customred} \textbf{dogs}} are \emph{daxable}, are {\color{customred} \textbf{corgis}} \emph{daxable}? 
    Here, \emph{daxable} is a nonce word used to study property inheritance without any confounding effects from parametric knowledge. 
    Language models may rely on taxonomic (left) and/or similarity (right) relations to perform property inheritance. %
     We investigate the interplay between these two effects in LMs' property inheritance judgments using both behavioral and mechanistic analyses.}
    \label{fig:das_diagram}
    \vspace{-1.5em}
\end{figure}

Debates about the mechanisms that underlie human property inheritance offer an exciting avenue to analyze conceptual organization and use in language models (LMs). 
First, due to the close connection of a word to its %
conceptual meaning \citep{murphy2004big, lupyan2019words, lake2021word}, property inheritance provides an opportunity to diagnose meaning-sensitivities in LMs \citep{piantadosi2022meaning, bender-koller-2020-climbing}.
Next, similarity effects like those found by \citet{sloman1998categorical} are related to the notion of \textit{content effects} shown by humans and LMs on logical reasoning tasks \citep{wason1968reasoning, evans1989bias, lampinen2024language}; the presence of similarity effects goes against the idea that humans rely on abstract taxonomic principles %
(where every category member would be equally likely to inherit the property). Finally, the principles of taxonomy vs.~similarity offer two clear hypotheses about what mechanisms could govern property inheritance behavior in a system.
This allows us to go beyond prior work investigating property inheritance in LMs \citep{misra-etal-2023-comps}, which has largely focused on their \textit{behavior}, leaving open  questions about the underlying mechanisms.

In this study, we empirically investigate the roles of taxonomic relations and categorical similarities on LMs' property inheritance behavior, using both behavioral and causal interpretability methods on four LMs. First, we behaviorally analyze LMs' sensitivity to taxonomic relations when performing property inheritance (\S\ref{sec:behavioral}). We demonstrate that noun similarity, known to correlate with human property inheritance behavior \citep{sloman1998categorical}, also correlates strongly with property inheritance judgments in language models, and explains many of the false positives and false negatives. Then, we causally localize property inheritance to specific %
activation subspaces (\S\ref{sec:causally-disentangling})
using distributed alignment search (DAS; \citealp{geigerfinding2024}). %
By training and evaluating DAS on controlled subsets of the data, we show that the subspaces responsible for property inheritance reflect both taxonomic and similarity features. 
Together, these results indicate that LMs are susceptible to non-trivial content effects rooted in noun similarities, and moreover that taxonomy and similarity are \emph{fundamentally entangled} in model representations.

\section{Conceptual Organization in LMs}

To what extent is the similarity vs.\ taxonomy debate relevant to LMs? By definition, LMs learn representations from distributional statistics of tokens in context, so
any reasoning they demonstrate is a result of distributional similarity. While technically true, this viewpoint
may be an oversimplification:
two words can be distributionally similar if they participate in antonymy (\textit{hot} vs. \textit{cold}), metonymy (\textit{car}, \textit{wheel}), hypernymy (\textit{robin}, \textit{bird}), or even if they share thematic relations (\textit{dog}, \textit{bone}). For property inheritance, however, it is only the \textbf{hypernymy} relation that plays a critical role \citep{murphy2004big}.
Is it possibile that something structural like a taxonomy might arise through distributional statistics? It has been postulated that symbolic behavior can emerge as a result of sub-symbolic processes internal to a neural network \citep{smolensky1986distributed, smolensky1988proper, mcclelland2010letting}. Empirically, this has been shown in a number of recent works, e.g., \citet{nanda2023progress} on LMs and modular arithmetic, and \citet{feng2023towards} on LMs learning to simulate dynamic programming---both of which require operating over symbolic representations.

The above discussion raises multiple interrelated questions: when performing property inheritance, do LMs show sensitivity to taxonomic relations? If so, is their behavior \emph{graded}, where properties are less likely to be projected to concepts that are taxonomically related to the higher-level category but low in similarity? Or, is it more \emph{binary}, where there is no significant difference between how likely a property is projected from the higher-level category to its high- and low-similarity members? If their behavior is graded, then does it differ between non-category members that might share similarities with the higher level category (e.g., bird and giraffe) vs.~those that do not (e.g., bird and chair)?

\paragraph{Related work} By focusing on the mechanisms that underlie LMs' property inheritance behavior, our work contributes to a rich body of work that investigates conceptual representations in language models \citep[][\textit{i.a.}]{bhatia2020transformer, misra2021typicality, patel2022mapping, abdou-etal-2021-language, grand2022semantic, wu-etal-2023-plms, park2024geometry}. Complementing these, our focus in this work is to understand how LMs \emph{use} their conceptual structure to project properties from one category to another---an important task in the broader space of inductive problems solved by humans \citep{kemp2014taxonomy}.

In this vein,
recent works have investigated LMs' property inheritance behavior---either by purely behavioral analyses \citep{misra-etal-2023-comps, wu-etal-2023-plms}, or by manipulating LMs' internal representations to edit their taxonomic knowledge \citep{powell-etal-2024-taxi, cohen-etal-2024-evaluating}, and measuring the implications of the edit. 
Importantly, these works assume that changes to taxonomic membership
must be directly reflected in the LMs' behavior, without necessarily considering the impact of categorical similarity. Grounding our investigation in empirical observations about humans \citep{sloman1998categorical}, we abstract away from this assumption, and explicitly consider the interplay between similarity and taxonomic relations by characterizing it using both behavioral and causal analysis methods.

\section{Experimental Materials}
\label{sec:expmats}

\subsection{Data} 
\label{sec:data}
We used the THINGS dataset \citep{hebart2019things, THINGSdata} as our primary repository of noun categories. THINGS is a %
repository of human responses on odd-one-out tasks for 1,854 unique object categories---e.g., \textit{dogs} and \textit{chairs}. Each object concept is also %
annotated with 
its superordinate category (\textit{dog}--\textit{animal}). 
We performed a few modifications to this dataset. First, many objects belonged to more than one category---e.g., \textit{cat} is an \textit{animal} but it is also a \textit{mammal}; we make these into separate entries. Second, we also paired each object with its WordNet sense, since one of the similarity methods we use involves ground-truth knowledge of word senses (we expand on this in \cref{sec:sim-scores}). We discarded instances for which we could not find a valid sense in WordNet. These manipulations result in 2,016 pairs of taxonomically related object-category pairs. Our final dataset has 44 superordinate and 1,281 subordinate categories, respectively. Next we describe our method to sample pairs that are not taxonomically related.

\paragraph{Negative Sampling}
Because we are explicitly analyzing the effect of \textit{similarity} in LMs' property inheritance behavior, we include similarity measures as a central component for deriving pairs of related categories that are \emph{not} taxonomically related. We perform this sampling as follows: for each superordinate category $C$, %
we construct the negative sample space $\mathcal{N}$ as the set of items that are \textit{outside} that category---e.g., for \textit{bird}, this could mean non-birds such as 
\textit{zebra, sofa,}
etc. We then compute the similarity of each item in $\mathcal{N}$ to the superordinate category using a given similarity measure. Finally, if $k$ is the number of category members of $C$, then we sample the top $\nicefrac{k}{2}$ and bottom $\nicefrac{k}{2}$ concepts from $\mathcal{N}$, based on the similarity values. This way, for each superordinate category, we have $k$ category members, and $k$ non-category members, with an even split between high and low-similarity non-taxonomic items.

\paragraph{Stimuli design}
We follow precedent from research in the psychology of concepts \citep{osherson1990category, sloman1998categorical}, and create stimuli in a premise-conclusion format, commonly used to analyze category-based inference in humans. In each instance, the premise expresses a statement where a category is said to have some property (\textit{birds have the T9 hormone}), and the conclusion asks if a different category also has that property (\textit{robins have the T9 hormone}). Following the same precedent, we use properties that are expressed by nonce words (e.g., \textit{are daxable}). 
Since LMs are likely to have less---if any---prior knowledge about these nonce properties, this design decision allows us to isolate LMs' inference behavior to the relations between categories (i.e., taxonomy and similarity), devoid from any interference from knowledge of real properties (e.g., \textit{can fly}) which may already be embedded in their learned weights.
We experimented with multiple different surface form realizations of the stimuli, 
and found the following template to work best on average:\footnote{Some models demonstrated better performance when we omitted ``\texttt{The answer is:}'' from this prompt. The entire set of templates %
is made available in Appendix \ref{sec:prompts}.}

\begin{displayquote}
    \small
    \texttt{Answer the question. Given that \textbf{A} is/are daxable, is it true that \textbf{B} is/are daxable? Answer with Yes/No. The answer is:}
\end{displayquote}
\noindent where \textbf{\texttt{A}} and \textbf{\texttt{B}} are the premise and conclusion categories, respectively.
Most nouns were pluralized, except for those which were more naturally expressed in singular form (mass nouns, e.g., \emph{honey}). A model that is sensitive to taxonomic relations when performing property inheritance should be more likely to generate \emph{Yes} than \emph{No} for a taxonomic pair such as (\textit{toy}, \textit{doll}),
while the reverse should be true for non-taxonomic pairs such as (\textit{fruit}, \textit{puppy}).

\begin{table*}[!t]
\centering
\resizebox{0.65\textwidth}{!}{%
\begin{tabular}{@{}lcll@{}}
\toprule
\textbf{Similarity Type} & \textbf{Category} & \textbf{Taxonomic Neighbors} & \textbf{Non-taxonomic Neighbors} \\ \midrule
Word-Sense & vehicle & \textit{\begin{tabular}[c]{@{}l@{}}car (0.85), sled (0.70), \\ hot-air balloon (0.59)\end{tabular}} & \textit{\begin{tabular}[c]{@{}l@{}}headlight (0.82), spinach (0.11), \\ calzone (0.09)\end{tabular}} \\ \midrule
SPOSE & bird & \textit{\begin{tabular}[c]{@{}l@{}}eagle (0.95), pelican (0.90), \\ penguin (0.82)\end{tabular}} & \textit{\begin{tabular}[c]{@{}l@{}}bee (0.92), gazelle (0.87), \\ cabinet (0.08)\end{tabular}} \\ \bottomrule
\end{tabular}%
}
\caption{Examples of taxonomically related and non-taxonomically related neighbors of the \textit{vehicle} and \textit{bird} categories determined by the two different types of similarities. Values in parentheses indicate similarities.}
\label{tab:sim-examples}
\vspace{-1em}
\end{table*}

\subsection{Similarity scores} 
\label{sec:sim-scores}
It is non-trivial to predict the nature of the `similarity' that might have an effect on a system's category-based inference behavior. This is true even for humans---most experiments used human derived similarity ratings to construct their stimuli \citep{osherson1990category, sloman1998categorical}, but what these ratings actually captured was largely unknown. To this end, we used two different types of similarity metrics, one sensitive to global distributional contexts of word-senses, and the other sensitive to the visual and conceptual properties of objects (\textit{animacy}, \textit{color}, etc.). Both metrics were derived from vector-space embeddings, and similarity was calculated using cosines between vectors of the input concepts.\footnote{We discuss other possible metrics we could have used, and our reasons for not using them in the limitations.} 

\paragraph{Word-sense Similarity} 
We used the LMMS-ALBERT-xxl model \citep{loureiro2022lmms} which was constructed by aggregating contextualized embeddings from the ALBERT-xxl LM \citep{lan2019albert} for lexical items with tagged WordNet senses. This model can be considered as a distributional semantic model like Glove \citep{pennington2014glove} that isolates the distributional information of the particular sense of a word---i.e., the embedding for \textit{bat} when used as a mammal is different from that used as sporting equipment.

\paragraph{\spose{} Similarity} 
We used \spose{} \citep{zheng2019revealing}, a sparse, non-negative vector space model fitted to human behavioral judgments on odd-one-out tasks, where participants were tasked to choose a concept from a triple %
that was least similar to the other two (e.g., $\{\textit{crow}, \textit{sparrow}, \textit{\textbf{ship}}\}$). We used the 49-dimensional embedding released by \citep{THINGSdata}, which achieved a correlation of up to 0.9 with human judgments of perceived similarity \citep{kaniuth2024high}. Since the embeddings in \spose{} are for subordinate categories, we augmented this dataset with superordinate category vectors by taking the mean of the vector representations of the category members for each category---e.g., for \textit{bird}, we took the average of the embeddings of $\{\textit{crow, sparrow, ...}\}$. %

\Cref{tab:sim-examples} shows examples of taxonomically and non-taxonomically related neighbors of a few categories according to both types of similarity.\footnote{Additional examples are given in Appendix~\ref{sec:concept-examples}.}
For each similarity type, we bin the category pairs in our stimuli into `high' or `low' similarity by calculating the median similarity for premise category and categorizing instances where the similarity between the premise and the conclusion is above the median to be `high', and those below to be `low'. This way, every premise category in our stimuli is associated with equal numbers of high and low similarity conclusion categories. In total, we have 4032 category pairs, for each type of similarity.

\begin{table*}[!t]
\centering
\resizebox{0.65\textwidth}{!}{%
\begin{tabular}{@{}lcccccccc@{}}
\toprule
\multirow{2}{*}{\textbf{Model}} & \multicolumn{4}{c}{\textbf{Word-Sense}} & \multicolumn{4}{c}{\textbf{SPoSE}} \\ \cmidrule(l){2-9} 
 & \textbf{TS} & \textbf{PS} & \textbf{MS} & \multicolumn{1}{c}{$\rho$} & \textbf{TS} & \textbf{PS} & \textbf{MS} & \multicolumn{1}{c}{$\rho$} \\ \midrule
Gemma-2-2B-IT & 0.84 & 0.85 & 0.89 & 0.26 & 0.79 & 0.79 & 0.88 & 0.59 \\
Mistral-7B-Instruct-v0.2 & 0.84 & 0.85 & 0.84 & \textbf{0.42} & 0.80 & 0.81 & 0.83 & 0.62 \\
Llama-3-8B-Instruct & 0.86 & 0.86 & \textbf{0.98} & 0.37 & 0.79 & 0.79 & \textbf{0.98} & \textbf{0.68} \\
Gemma-2-9B-IT & \textbf{0.90} & \textbf{0.89} & 0.85 & 0.34 & \textbf{0.85} & \textbf{0.84} & 0.84 & 0.65 \\ \bottomrule
\end{tabular}%
}
\caption{Behavioral sensitivities of LMs across both similarity types (Word-Sense and \spose{}). TS: Taxonomic Sensitivity; PS: Property Sensitivity; MS: Mismatch Sensitivity; $\rho$: Spearman correlation of model scores ($P_{\texttt{rel}}(\text{Yes})$) with similarity ($p < .001$ throughout).}
\label{tab:behavioral-sensitivities}
\vspace{-1em}
\end{table*}

\subsection{Models}
\label{sec:models}
We analyze four instruction-tuned LMs: Mistral 7B Instruct v0.2 \citep{jiang2023mistral}, the 2B and the 9B parameter variants of the Gemma 2 family \citep{riviere2024gemma2}, and the 8B parameter variant of Llama 3 Instruct \citep{llama3}.\footnote{Initial experiments showed these LMs' non instruction-tuned counterparts to perform worse at property inheritance, which corroborates recent evidence about the advantage of instruction-tuning for property inheritance \citep{misra-etal-2024-experimental}.} We found varying degrees of success in using the chat-templates for these LMs, and therefore used it only when it resulted in a non-trivial improvement over the standard method of passing inputs to the LMs. All models were accessed using the \texttt{transformers} library \citep{wolf-etal-2020-transformers}.

\subsection{Behavioral Characterization of Property Inheritance}\label{sec:behavioral}

We begin by characterizing the role of taxonomic relations and similarity in LMs' property inheritance behavior. To this end, we collect LMs' responses to our stimuli, and analyze the conditions under which they are more likely to extend the property from the premise category to the conclusion category. We do this by comparing their relative probabilities for `Yes' vs. `No' on our stimuli, which we compute  as follows:\footnote{The LMs that we use have two different tokens to represent `Yes'---one with and one without space, and similarly for `No'. To address this potential issue, we take the maximum of the probabilities over the two options for both cases. That is, $p_\theta(\text{Yes}) = \max(p_\theta(\texttt{Yes}), p_\theta(\texttt{\textvisiblespace Yes}))$, and similarly for `No'. We note that these tokens appeared as the top predicted tokens for all models evaluated except Gemma-2-2B-IT.}
\begin{align*}
    P_{\texttt{rel}}(x) = \frac{p_\theta(x \mid \text{prefix})}{\sum\limits_{x \in \{\text{Yes, No}\}}p_\theta(x \mid \text{prefix})},
\end{align*}
\noindent
where `prefix' is the stimulus, formatted as discussed in \Cref{sec:expmats}, and $p_\theta(.)$ denotes the LM's next token %
distribution which we computed using the \texttt{minicons} library \citep{misra2022minicons}. Using these relative probabilities, we compute the following metrics to analyze LMs' property inheritance behavior:

\paragraph{Taxonomic Sensitivity (TS)} We measure the extent to which LMs are sensitive to the taxonomic relations between the premise and conclusion category. We do so by computing the proportion of examples where $P_{\texttt{rel}}(\text{Yes}) > 0.5$ when the premise and conclusion are taxonomically related, and  $P_{\texttt{rel}}(\text{Yes}) < 0.5$ when they are not. 

\paragraph{Property Sensitivity (PS)} 
The TS metric is computed for stimuli where the only nonce property is \textit{is/are daxable}. To what extent does the specific surface form of the property play a role in LMs' property inheritance behavior? To this end, we generate a different set of stimuli where we randomly replace both instances of \textit{is/are daxable} with \textit{has/have feps} for half of the examples. We maintain the even balance between taxonomically and non-taxonomically related premise-conclusion pairs. Given this data, we then compute the taxonomic sensitivity in this setting as before. This metric penalizes bias for a particular property.

\paragraph{Mismatch Sensitivity (MS)} 
We measure the extent to which the LMs are sensitive to whether the properties mentioned in the premise and conclusion match. If they are mismatched, then regardless of the taxonomic relationship between the concepts, the LMs should always prefer `No' over `Yes'. To test this, we generate stimuli where we uniformly substitute either the premise or the conclusion with a different property (e.g., \textit{birds are daxable} vs.\ \textit{robins have feps}), and compute the proportion of time $P_{\texttt{rel}}(\text{Yes}) < 0.5$.

\begin{figure*}[h]
    \centering
    \includegraphics[width=0.85\linewidth]{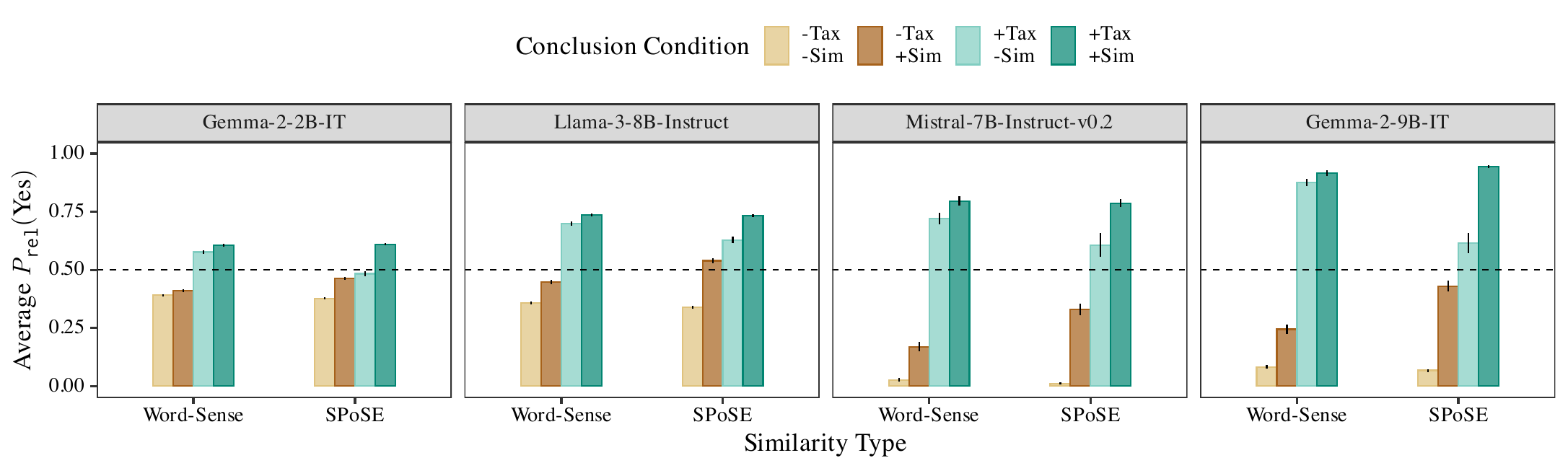}
    \caption{LMs' Average Relative Probability of `Yes' for different conclusion categories and different types of similarities (Word-Sense vs SPoSE). LMs show a clear sensitivity to taxonomic relations, but also show an effect of similarity, where they are more likely to extend the property to a conclusion category when the premise and the conclusion categories are highly similar. Chance behavior is 0.50, as indicated by the dashed line.}
    \label{fig:behavioral-results}
    \vspace{-1em}
\end{figure*}

\paragraph{Spearman correlation with similarity ($\rho$)} Finally, we measure the Spearman correlation between $P_{\texttt{rel}}(\text{Yes})$ and the similarity between the premise and conclusion categories, to quantify the role of similarity on LMs' inference behavior.

We compute these metrics separately for both types of similarity. For the three sensitivity metrics (TS, PS, MS), chance sensitivity is 0.5, as they all involve pairwise comparisons. We also measure the average $P_{\texttt{rel}}(\text{Yes})$ across four slices of our data based on when the premise and conclusion categories were (i) non-taxonomically related, low similarity (-Tax, -Sim), (ii) non-taxonomically related, high similarity (-Tax, +Sim), (iii) taxonomically related, low similarity (+Tax, -Sim), and (iv) taxonomically related, high similarity (+Tax, +Sim).

\subsection{Results and Analysis}
\Cref{tab:behavioral-sensitivities} shows the aforementioned metrics %
for all four LMs, across both similarity types, while \Cref{fig:behavioral-results} shows their average $P_{\texttt{rel}}(\text{Yes})$ values across the four different slices of our data.
For both similarity types, all four LMs demonstrate substantially high TS and PS values---i.e., they are more likely to extend the novel property when the premise category is a superordinate of the conclusion category. This suggests a non-trivial role of taxonomic category membership in the models' inference behavior. At the same time, the LMs also show positive correlation with categorical similarity, suggesting that their tendency to extend properties from the premise to the conclusion is also sensitive to the similarity between the two. More specifically, they show greater correspondence with similarities derived from \spose{} ($\rho \in [0.59, 0.68]$) than from the sense embeddings ($\rho \in [0.26, 0.42]$). 

While LMs %
demonstrate sensitivity to similarity, their tendency to produce `Yes' over `No' seems to correspond better with the presence of taxonomic relations. In \Cref{fig:behavioral-results}, LMs on average prefer to extend the property from premise to conclusion for high-similarity pairs than for low-similarity ones. However, we see the average $P_{\texttt{rel}}(\text{Yes})$ to be largely greater than 0.5 for the slices of our data where the premise is a taxonomic superordinate of the conclusion, regardless of the similarity bin (high vs.~low), for both types of similarity. For non-taxonomically related premise conclusion pairs, while the LMs show greater $P_{\texttt{rel}}(\text{Yes})$ value when the premise and conclusion are highly similar, it is seldom above 0.5 (with the exception of Llama-3-8B-Instruct, for the \spose{} similarity). 
Finally, all LMs show robustness to cases where the property being projected between the premise and conclusions was different (largely high MS values). That is, whatever taxonomic/similarity sensitivities LMs demonstrate are largely for the well-formed instances of the task (i.e., when the properties between premise and conclusion match).
These results suggest that \textbf{LMs' property inheritance behavior can be explained jointly by taxonomic relations as well as categorical similarity}.

\begin{figure*}[!t]
    \centering
    \includegraphics[trim=0 350 0 0, clip, width=.75\textwidth]{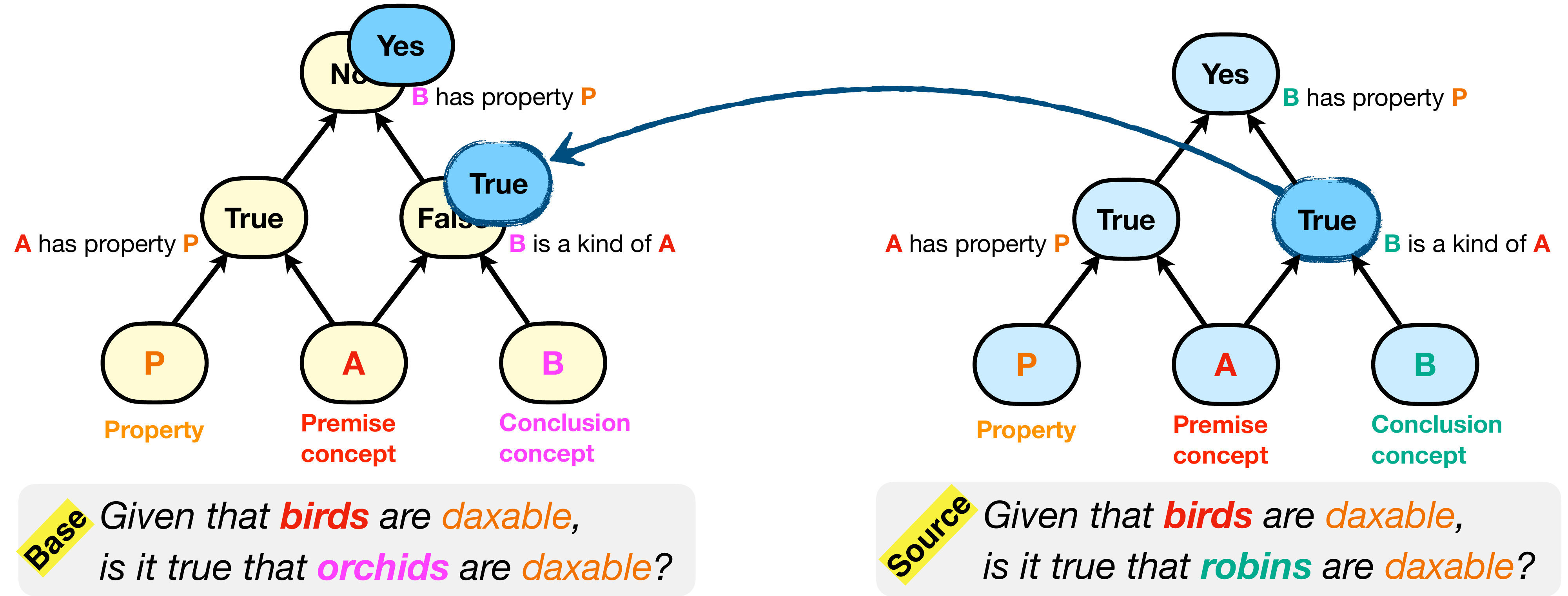}
    \caption{We hypothesize that language models rely on this causal graph to perform the categorical inference task. We are interested specifically in the node responsible for taxonomic judgments, so we use DAS (\S\ref{sec:causally-disentangling}) to isolate the subspace encoding this causal variable. We causally verify its sensitivity to taxonomy by setting its value to what it would have been on counterfactual \emph{source} inputs, and observing whether model behavior changes appropriately. Here, activations at the isolated subspace for the \emph{base} are replaced with activations from the \emph{source}. IIA measures to what extent this intervention results in the expected behavior given the hypothesized causal graph across inputs.}
    \label{fig:das_diagram}
    \vspace{-1em}
\end{figure*}
\section{Causal Investigation of Taxonomy and Similarity}
\label{sec:causally-disentangling}

Given that LMs perform property inheritance with high taxonomic sensitivity, we now use causal interpretability tools \citep{mueller2024quest,geiger2024causalabstractiontheoreticalfoundation}
to localize this behavior. %
We then investigate (i) whether the subspaces that are causally responsible for inheritance perform inductive generalization in a manner that is better explained by the similarity of the noun concepts, rather than their taxonomic relationship; and (ii) whether taxonomic information and noun similarity are fundamentally entangled in the model representations. %
We investigate both questions using \textbf{distributed alignment search} (DAS; \citealp{geigerfinding2024}), a causal interpretability method which has been used to localize and mechanistically explain syntax-sensitive behaviors \citep{arora-etal-2024-causalgym} and arithmetic abilities \citep{wu2023interpretability} in LMs. %
To avoid needing to search over all possible subspaces of an activation vector manually, we employ the Boundless DAS variant \citep{wu2023interpretability} implemented in \texttt{pyvene} \citep{wu-etal-2024-pyvene} to learn the appropriate subspace.

We use DAS rather than other causal interpretability techniques\footnote{For example, those based on sparse autoencoders \citep{bricken2023monosemanticity, marks2024sparse, huben2024sparse} or neurons \citep{vig2020investigating, finlayson-etal-2021-causal}.} because it allows us to investigate to what extent a specific hypothesized \emph{causal model} is being implemented by the neural network. 
Because DAS isolates a specific %
subspace corresponding to a given hypothesized causal variable, the subspace can be evaluated in novel contexts post hoc to better characterize its functional role. We leverage this to evaluate whether subspaces that are sensitive to \emph{taxonomic} relationships are also sensitive to \emph{similarity-based but non-taxonomic} relationships. 
This type of experimental design involving ambiguous training sets with separate test and generalization sets has been used to characterize the inductive biases of LMs \citep{mccoy-etal-2019-right,
si-etal-2023-measuring, mueller-etal-2024-context}, though this has typically been evaluated behaviorally rather than at the level of internal mechanisms. There is a limited but growing literature validating how well discovered mechanisms (or explanations thereof) generalize to novel types of contexts \citep[][\textit{i.a.}]{geiger-etal-2020-neural,huang-etal-2023-rigorously,huang-etal-2024-ravel}.

Given an LM $M$, a hypothesized causal graph $L$, and counterfactual input (source, base) pairs $(s,b)$ that target a specific intermediate variable $V$ in the causal graph, DAS learns a rotation $R$ 
for the activations of a given submodule, and then performs counterfactual interventions in this rotated space. This is done as follows: $L_V(s)$, the value of variable $V$ in $L$ when applied to source $s$, is patched into the corresponding variable of the base $b$,
\[
  L_V(b) \leftarrow L_V(s),
\]
and then the final output of $L$ is computed.
Translating this to operations on the low-level computation graph, a rotation $R$ is learned for activations in the network $M$ from a position $i$ to position $j$. We patch $R(M(s)_i)$ into $M(b)_j$:
\[
  M(b)_j \leftarrow R\bigl(M(s)_i\bigr),
\]
and then compute the output of $M$.

After learning the rotation,  we compute the interchange intervention accuracy (IIA) on a held-out test set. This scores how frequently the learned subspace performs as expected given the hypothesized causal graph. Intuitively, IIA is defined as the proportion of $(s,b)$ pairs where patching in the rotated source activation into the base model $M$ causes $M$'s output to match the causal model $L$'s output \emph{after} the intervention. It thus measures the degree of alignment between the neural network and the causal graph. The intuition for learning a rotation is that the features of interest may not necessarily be aligned to the bases %
of the original activation space \citep{hinton1986distributed,smolensky1986distributed,elhage2022toymodelssuperposition,mueller2024quest}, %
rendering neuron-based approaches insufficient for capturing the target variable. %
We refer readers to \citet{geigerfinding2024} for details.

\begin{figure*}[!t]
    \centering
    \includegraphics[width=0.9\linewidth]{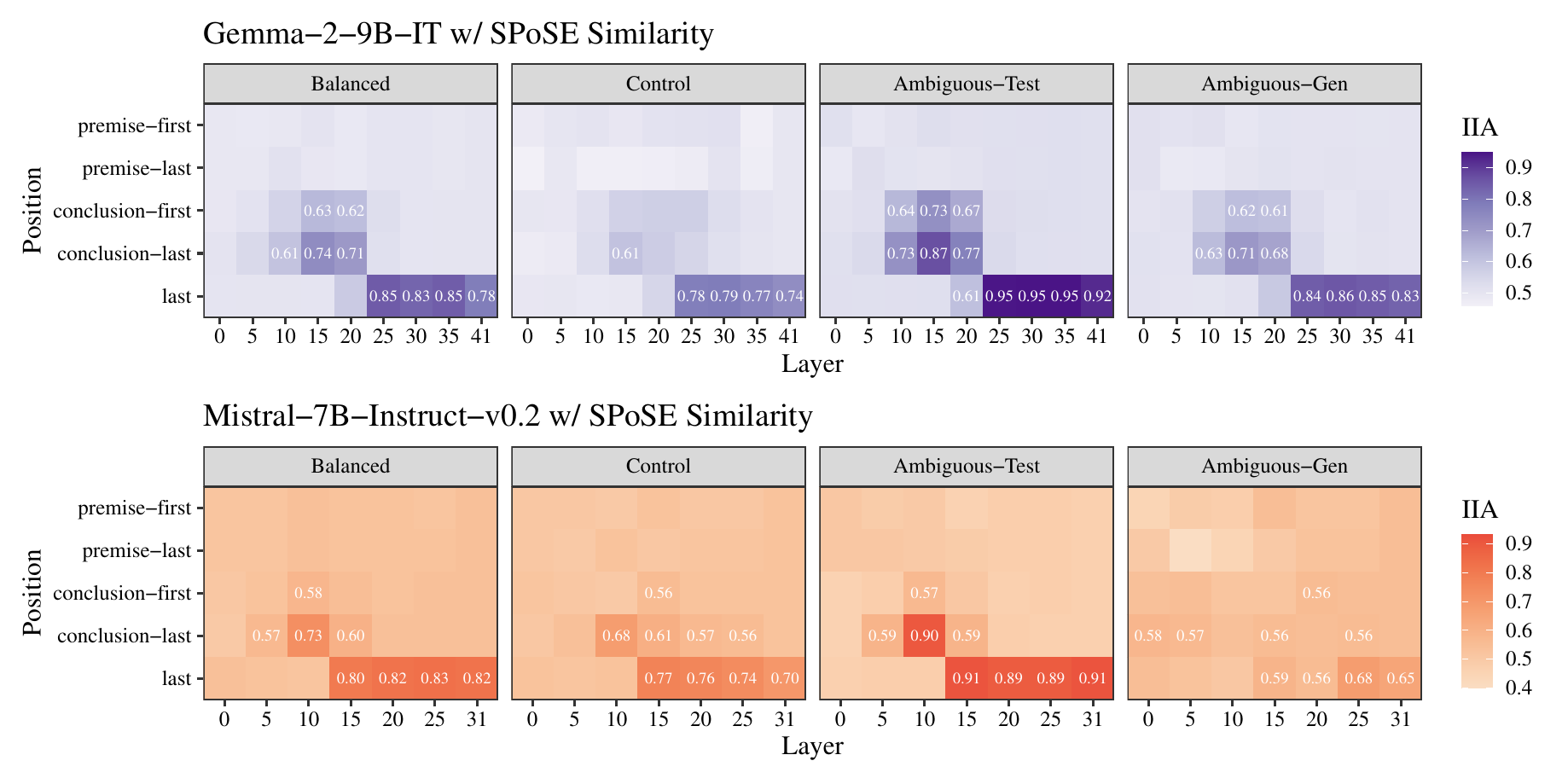}
    \caption{Interchange Intervention Accuracies (IIA) for the causal graph in \Cref{fig:das_diagram} for \textbf{Gemma-2-9B-IT} (Top) and \textbf{Mistral-7B-Instruct-0.2} (Bottom), when intervening at various layers and token positions, with negative samples derived using \spose{} similarities. Since the premise and conclusion nouns are often multiple tokens, we show IIA when intervening at the first and last positions of each. \textbf{Note:} Both models have different numbers of layers.}
    \label{fig:gemma-mistral-das}
    \vspace{-1em}
\end{figure*}

In our experiments, we hypothesize that a given subspace will be sensitive to taxonomic relationships---if this is true, then passing an activation into this subspace from taxonomically unrelated inputs should yield a variable output of \texttt{False}. However, if we intervene on this subspace passing in activations from an alternate input where the nouns \emph{are} taxonomically related (while holding all other activations in the model constant), then the variable's output should flip to \texttt{True}. We illustrate the intuition and the causal graph that we hypothesize LMs implement in \Cref{fig:das_diagram}. %
By only changing the premise and conclusion nouns across inputs (e.g., \emph{birds}, \emph{robins}, \emph{orchids}), and keeping the rest of input constant, we effectively target the node ``\textbf{\textcolor{customgreen}{B}} is a kind of \textbf{\textcolor{red}{A}}'' in \Cref{fig:das_diagram}.\footnote{We target this node since our focus is on taxonomic relations rather than property binding---where the goal is to check if LMs have appropriately bound the property with the noun concept---which has been investigated in other work \citep{feng2024how}.}

We train DAS %
on a subset of 3,000 stimuli and evaluate on the rest. Implementation and training details are given in Appendix~\ref{sec:das-details}. Next, we describe variations of the training and test sets used in our experiments.

\paragraph{Balanced} 
For each \emph{base} stimulus in the set of 3,000 stimuli, we pick a \emph{source} counterfactual stimulus by sampling from this set of stimuli without replacement. This ensures a balanced coverage of counterfactual stimuli in terms of similarity and whether or not the premise and conclusion are taxonomically related. This experimental setting is used to localize property inheritance in the network.

\paragraph{Control} DAS requires training, raising concerns that its expressivity may lead to discovered causal effects where none exist \citep{arora-etal-2024-causalgym}, similar to concerns raised in the probing literature \citep{hewitt-liang-2019-designing}. To mitigate this, we follow \citet{arora-etal-2024-causalgym} and map the labels \texttt{Yes} and \texttt{No} to the semantically irrelevant words \texttt{chart} and \texttt{view}, respectively. This setting evaluates the extent to which DAS memorizes an arbitrary token mapping versus learning task-specific structure.%

\paragraph{Ambiguous} In this setting, we filter the training set such that learned rotations can be consistent with \textit{either} similarity or taxonomic features. By only training on taxonomic examples with high similarity and negative examples with low similarity, we learn an intervention which is ambiguous with respect to what features it is using. To disambiguate these, we evaluate on two different test sets: an in-domain (\textbf{Ambiguous-Test}) test set composed in the same way as the train set (taxonomic examples with high similarity and negative examples with low similarity), and an out-of-domain generalization test set (\textbf{Ambiguous-Gen}), composed of  taxonomic examples with low similarity and non-taxonomic examples with high similarity. 
A drop in IIA values between the Test and Generalization settings would show that learned interventions rely more on similarity than taxonomic features.%
\paragraph{Unambiguous} 
Finally, we test whether the subspace sensitive to taxonomic relationships is also sensitive to similarity-based relationships---i.e., whether taxonomic information and noun similarity are entangled in the model. To do this, we take the intervention trained under the \textbf{Balanced} setting and evaluate it on (\textbf{Ambiguous-Gen}), defined above. Here the training signal for DAS is unambigious--it is only compatible with capturing taxonomic information. A drop in scores would then indicate that taxonomic and similarity features are entangled.

We use $P_{\texttt{rel}}$(Yes) rather than the top predicted token as done in \citep{wu2023interpretability} when computing IIA across experiments, in order to have a reasonable comparison against the control setting.\footnote{The interventions are not strong enough to force the model to output either of the control tokens \texttt{chart}, or \texttt{view} as their top-predicted token, resulting in 0\% accuracy; on the other hand the difference between using the top predicted token and $P_{\texttt{rel}}$(Yes) was negligible in most other settings (<0.02 across experiments, except for Gemma-2-2B-IT).}

\subsection{Results and Analysis}

\Cref{fig:gemma-mistral-das} shows the IIA results for Gemma-2-9B-IT and Mistral-7B-Instruct-v0.2 for the first three settings, and \Cref{tab:max-iia-values} shows results on the \textbf{Unambiguous} setting. Full results for the other models are given in Appendix \ref{sec:additional_results}. Higher IIA values indicate a better causal alignment between the intervention at a particular location in the network and the causal graph. The activations at the premise token positions have poor alignment with the causal graph, %
while activations around the last conclusion token at layer 10 (for Mistral) and layers 10-20 (for Gemma) have a much bigger causal effect on property inheritance.

IIA values for the control experiment are always lower than for the balanced setting; thus, despite its expressivity, DAS can yield task-dependent results (i.e., comparatively lower IIA values in the control setting establishes that DAS is telling us something about the LM’s property inheritance abilities).
High values for IIA occur in roughly the same positions for Ambiguous-Test and Ambiguous-Gen. There is a substantial drop from Ambigious-Test to Ambigious-Gen for all models and for every intervention position, showing that, \textbf{when given ambiguous data, the learned rotations tend to rely on similarity rather than taxonomic relations.} 
Similar observations hold for the Unambigious experiment, with a drop in scores showing that, even when trained on data that is unambigiously consistent with taxonomic relations rather than similarity, the causal model is sensitive to similarity effects. In other words, \textbf{similarity and taxonomic features appear to be fundamentally entangled}.

\section{Discussion and Conclusions}
Our goal in this paper was to characterize category-based inferences in LMs.~We derived inspiration from psychological experiments that have sought to understand how humans transfer properties from a higher level category like \textit{mammals} to lower level ones such \textit{dogs} and \textit{cats} \citep{collins1969retrieval, glass1974alternative, murphy2004big}---i.e., show \textit{property inheritance}. Findings from these experiments show that instead of relying purely on abstract category-membership principles (where all members of a category are equally likely to inherit properties), humans show graded behavior 
proportional to the similarity between the categories \cite{sloman1998categorical}.

\begin{table}[!t]
\centering
\resizebox{0.45\textwidth}{!}{%
\begin{tabular}{@{}lcccc@{}}
\toprule
\multirow{2}{*}{\textbf{Model}} & \multicolumn{2}{c}{\textbf{Word-Sense}} & \multicolumn{2}{c}{\textbf{SPoSE}} \\ \cmidrule(l){2-5} 
 & \textbf{Bal} & \multicolumn{1}{c}{\textbf{Gen}} & \textbf{Bal} & \multicolumn{1}{c}{\textbf{Gen}} \\ \midrule
Gemma-2-2B-IT & 0.86 & 0.81 & 0.80 & 0.63 \\
Mistral-7B-Instruct-v0.2  & 0.88 & 0.85 & 0.83 & 0.65 \\
Llama-3-8B-Instruct  & 0.87 & 0.82 & 0.80 & 0.63 \\
Gemma-2-9B-IT  & 0.90 & 0.85 & 0.85 & 0.67 \\ \bottomrule
\end{tabular}%
}
\caption{Maximum IIA values for the Unambiguous experiment. Interventions are trained under the Balanced train set, and evaluated on both the Balanced test set (Bal) as well as the Ambiguous-Gen set (Gen). The drop on the generalization test set suggests that the learned interventions rely non-trivially on similarity.}
\label{tab:max-iia-values}
\vspace{-1em}
\end{table}

Through behavioral and representational analyses methods across four LMs, we found that \textbf{language models' property inheritance behavior is sensitive to both taxonomic relations \emph{and} similarity}.~Behaviorally, LMs were more likely to extend the property from the premise to the conclusion when they were taxonomically related, and at the same time, showed positive correlations with the similarity between them. This finding was further reinforced in our representational investigations, where \textbf{we found taxonomy and similarity to be largely entangled in the LM subspaces causally responsible for property inheritance.} 
Our results cast doubt on previous works' assumption of pure taxonomic principles with respect to property inheritance in LMs, without considering similarity \citep{misra-etal-2023-comps, powell-etal-2024-taxi, wang2024editing}.
In the broader context of LMs' reasoning behaviors, our results showcase yet another instance where LMs were susceptible to human-like content effects \citep{lampinen2024language}.

Our findings also suggest interesting hypotheses for future human analyses.~Importantly, \citet{sloman1998categorical}'s experiments never tested humans on cases where the categories in a property inheritance setting did \textit{not} share taxonomic relations (e.g., birds and bats). This makes it unclear if the principles of taxonomy and similarity are \textit{also} not mutually exclusive for humans \citep{murphy2012semantic}---e.g., it could be possible that both mechanisms are at play, where taxonomic relations demarcate the inference (i.e., whether the property will be inherited) and similarity modulates its strength.
Future experiments could test this hypothesis and shed further light on the extent to which LMs and humans show similar sensitivities and biases during reasoning. Overall, our work shows how tools from modern interpretability research, primarily used to isolate sub-structures in neural networks, may also be used to perform hypothesis testing of the abstractions encoded within them.%

\section*{Limitations}

\paragraph{Alternative similarities} 
There are other similarities that we could have used in our experiments. Examples include similarities computed over WordNet (e.g., Wu-Palmer similarity; \citealp{wu-palmer-1994-verb}), co-occurrence based similarities, or similarities computed from LM token embeddings or activations \citep{chronis-erk-2020-bishop}. %
While it is tempting to use these, there are various shortcomings to each of them. WordNet similarities do not distinguish between co-hyponyms: concepts that are at the same level will have the same similarity with a given high level concept. Next, the sense based similarity used in this work \textit{is} a form of distributional similarity as it is formed by aggregating distributional semantic embeddings of words occurring in free-text corpora, and has the added advantage of being informed by the sense of the word. Finally, using LM-specific similarities makes it difficult to use the same stimuli across models, since similarities from different models would have resulted in different negative samples. However, future work could explore possible correlations between LM-derived similarities and their property inheritance behavior.

\paragraph{Abstract and Ad-hoc Concepts} 
We have exclusively focused on concrete, object noun concepts in this work.
However, these do not account for the entire range of concepts that humans acquire in their life times---i.e., it is unclear how property inheritance applies to abstract concepts like \textit{love} or ad-hoc concepts \citep{barsalou1983ad} like \textit{things you would pack for a camping trip}, which may not have taxonomic structure. We leave these investigations to future work.

\paragraph{Knowledge editing} Previous work has studied whether language models are able to carry out inferences related to entities \citep{cohen-etal-2024-evaluating} or concepts \citep{wang2024editing,powell-etal-2024-taxi} which have been modified through knowledge editing. While we do not engage directly with knowledge edits, it would be interesting to measure the impact of knowledge editing on the LMs' rotated subspaces during property inheritance.

\paragraph{Broader space of inductive problems} While property inheritance is an important consequence of category structure and organization, it is only one type of inductive generalization that humans exhibit \citep{kemp2014taxonomy}. For instance, humans may also learn about novel concepts and ascribe known features to them \citep{smith1978theories, murphy2004big, kemp2011inductive}, or they might learn a new higher level category and judge category memberships of known concepts \citep{xu2007word}, or they might generalize in a completely non-deductive manner (e.g., from robins to birds, or to its co-hyponyms, see \citealp{rips1975inductive, osherson1990category}). 

\paragraph{Contextual similarities} While we experimented with different types of similarities, human inductive inferences might not be explained by the same similarity every time---i.e., similarity in inductive problems can be context-sensitive \citep{heit1994similarity, rogers2004semantic, kemp2009structured}. For instance, generalization of biological-sounding properties (\textit{has an ulnar artery}) are often sensitive to taxonomic similarities. Bears and whales are likely to share these properties since they are mammals. On the other hand, behavioral properties (\textit{studies its food before attacking}) might be influenced by similarities derived from other, non-taxonomic conceptual structures. The aforementioned property could be shared by predatory animals such as tigers and eagles. 
It is, however, important to note that these caveats apply only when the properties are not \textit{nonce}, like in our work---e.g., it is not clear if \textit{is daxable} or \textit{has feps} are biological or otherwise. Future studies could explore these deeper nuances of similarity and category-based inferences in LMs using the methods we employ in this work.

\section*{Acknowledgments}
A.M. is supported by a postdoctoral fellowship under the Zuckerman STEM Leadership program. 
K.M. acknowledges postdoctoral funding supported by NSF Grant 2139005 awarded to Kyle Mahowald. We also thank Martin Hebart for help on the THINGS dataset and the SPoSE embeddings, Atticus Geiger and Zhengxuan Wu for assistance with \texttt{pyvene}, and Katrin Erk and the anonymous reviewers for helpful discussion and feedback.

\bibliography{custom, kanishka}
\appendix

\section{Sensitivity to directionality}
\label{sec:more-details-directional}

Our previous analyses suggest that LMs demonstrate non-trivial taxonomic sensitivity in their property inheritance behavior. Are they also sensitive to the fact that the taxonomic relation is, in its strictest sense, asymmetrical \citep{murphy2004big}? For instance, while \textit{robins} are \textit{birds}, one cannot say that \textit{birds} are \textit{robins}. Is this fact captured in LMs' property projections? That is, if \textit{robins} are daxable, are \textit{birds} daxable? A finding of insensitivity might not necessarily be a negative result in the context of LMs' conceptual representations. 
After all, generalizing from subordinates (\textit{robins}) to superordinates (\textit{birds}) is directly connected to contemporary work targeting property induction in LMs \citep{misra2022property, han2022human, bhatia2023inductive, han2024inductive}, where similarity has more of an uncontroversial role \citep{osherson1990category, sloman1993feature}, and where humans show a myriad of interesting phenomena \citep{hayes2018inductive}.\footnote{We refer to this analysis %
as the \emph{directionality} of property projection given that \emph{induction} is often used more generally to refer to many kinds of reasoning which are \textit{ampliative}, and not determined by deductive logic \citep{kemp2014taxonomy}.}
Instead, it might be informative to further characterize the LMs' subspaces responsible for property inheritance---our previous findings suggest that they are sensitive to similarity, are they also sensitive to directionality? Or are they simply encoding the presence of taxonomic (and similarity) relations?

We behaviorally evaluate the sensitivity of LMs to the direction of the property projection as follows. We measure the \textbf{Directional Sensitivity} (DS) as the fraction of times an LM flips its relative preference from \texttt{Yes} to \texttt{No} when the order of the premise and conclusion nouns in our stimuli is swapped, \textit{only} over all taxonomically related items. Additionally, we also measure the spearman correlation of DS with that of TS---i.e., the taxonomic sensitivity metric computed for the non-reversed property inheritance stimuli, computed for the taxonomically related premise and conclusion. We refer to this as $\rho_{\textrm{DS}}$

\begin{figure}[t]
    \centering
    \includegraphics[trim=0 10 0 0, clip, width=1\linewidth]{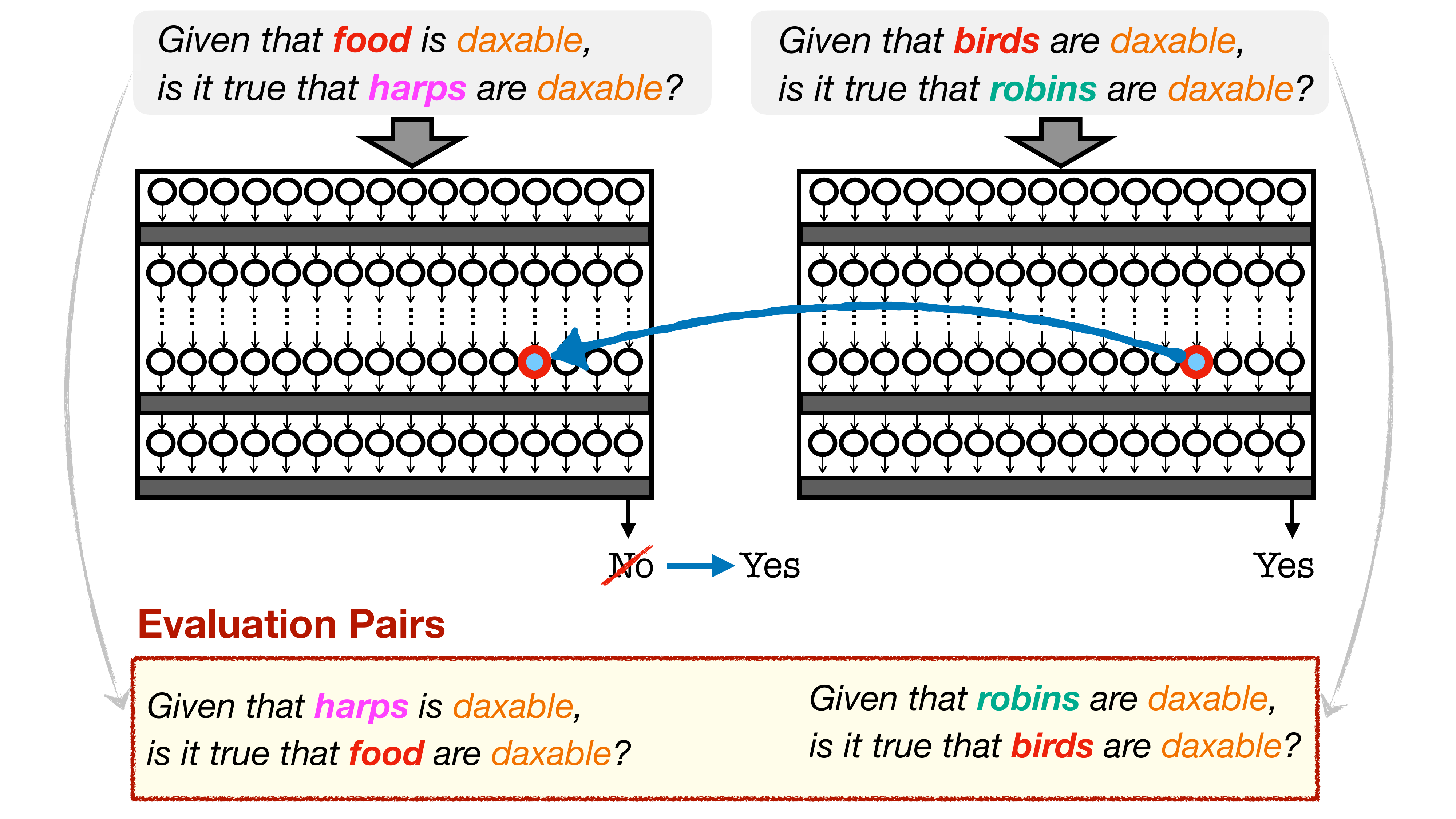}
    \caption{DAS interventions on pairs where the direction of the inference is flipped. Low IIA values for this experiment reveal whether the learned subspaces for property inheritance are sensitive to direction.}
    \label{fig:directionality_figure}
    \vspace{-1em}
\end{figure}

Next we consider whether the learned DAS interventions from \Cref{sec:causally-disentangling} generalize when reversing the direction of the premises and conclusions. We consider interventions trained under the \textbf{Balanced} setting with high IIA values (i.e., which localized property projection behavior in the regular property inheritance direction and showed taxonomic sensitivity), and evaluate them on source and base pairs with flipped premise and conclusion nouns.
In order to measure the effect of direction, we only evaluate on examples where the intervention successfully flipped the label from \texttt{No} to \texttt{Yes} in the non-reversed  direction, as illustrated in \Cref{fig:directionality_figure}. 
The evaluation set is constructed from source and base pairs where the premise and conclusions are taxonomic and non-taxonomic, respectively. We then filtered these so that (i) the LM predicts the right labels in the non-reversed case (i.e., shows taxonomic sensitivity) and (ii) the DAS intervention correctly flips the label from \texttt{No} to \texttt{Yes}. 
The order of the nouns are then reversed to establish whether the intervention causes the same behavior when the direction of inheritance is flipped. This is illustrated in \Cref{fig:directionality_figure}.

We refer to the IIA values for this evaluation as \textbf{Subspace Directional Insensitivity} (SDI), since higher scores mean the subspace is \emph{less} sensitive to direction.

\paragraph{Results}

The behavioral DS results across models are given in Table~\ref{tab:directional-sensitivities}. %
Gemma-2-9B-IT is the most directionally sensitive model: the order of the premise and conclusion concepts has an impact on property projection roughly half of the time.
The other three models are far less sensitive to direction. This is also reflected in higher Spearman correlations for those models.

Subspace Directional Insensitivity results are shown in Table~\ref{tab:das-directional-sensitivity} for interventions at different network locations. Since IIA values were high at both the final token position (Last) and the last token of the conclusion noun (Con-$\ell$), we evaluated with interventions at both, in each case selecting the layer with highest IIA across each token position.\footnote{For Gemma-2-9B-IT, this was layer 15 for Con-$\ell$, and layer 35 for Last; for the other models this was layer 10 for Conl-$\ell$ and layer 15 for Last.}

SDI is nearly always lower for Con-$\ell$ than for Last positions: it is harder to use the taxonomically sensitive property inheritance intervention to make the LM behave in the same way with reversed examples when intervening earlier in the network. Similar to the behavioral results, the subspace at conclusion-last token position and layer 15 for Gemma-2-9B-IT is the most sensitive to the directionality of the inference, while Gemma-2-2B-IT is the least sensitive. We note, however, that this only holds for the prompts used in our experiments---for example, using \emph{Prompt 2} with chat template (Appendix \ref{sec:prompts}) makes Gemma-2-2B-IT more directionally sensitive than it is with \emph{Prompt 1} without chat template.\footnote{Specifically, DS increases from 0.24 to 0.55, and SDI decreases (e.g., for Con-$\ell$, to 0.22 and 0.27 for Word-Sense and SPoSE, respectively). Future work could systematically explore direction sensitivity for different prompts.}

\begin{table}[t]
\centering
\resizebox{0.3\textwidth}{!}{%
\begin{tabular}{@{}lcc@{}}
\toprule
\textbf{Model} & \textbf{DS} & $\rho_{\text{DS}}$ \\ \midrule
Gemma-2-2B-IT & 0.24 & \textbf{0.54} \\
Mistral-7B-Instruct-v0.2 & 0.37 & 0.37 \\
Llama-3-8B-Instruct & 0.38 & 0.40 \\
Gemma-2-9B-IT & \textbf{0.49} & 0.31 \\ \bottomrule
\end{tabular}%
}
\caption{Directional sensitivities of LMs. DS: Directional Sensitivity; $\rho$: Spearman correlation of $P_{\texttt{rel}}$(Yes) between the original and reverse directions. DS and $\rho$ are computed only over the taxonomic items.}
\label{tab:directional-sensitivities}
\end{table}

\begin{table}[!t]
\centering
\resizebox{0.45\textwidth}{!}{%
\begin{tabular}{@{}lcccc@{}}
\toprule
\multirow{2}{*}{\textbf{Model}} & \multicolumn{2}{c}{\textbf{Word-Sense}} & \multicolumn{2}{c}{\textbf{SPoSE}} \\ \cmidrule(l){2-5} 
 & \textbf{Con-$\ell$} & \multicolumn{1}{c}{\textbf{Last}} & \textbf{Con-$\ell$} & \multicolumn{1}{c}{\textbf{Last}} \\ \midrule
Gemma-2-2B-IT & 0.71 & 0.77 & 0.77 & 0.83 \\
Mistral-7B-Instruct-v0.2 & 0.40 & \textbf{0.31} & 0.29 & \textbf{0.38} \\
Llama-3-8B-Instruct & \textbf{0.06} & 0.52 & 0.16 & 0.45 \\
Gemma-2-9B-IT & 0.08 & 0.57 & \textbf{0.07} & 0.39 \\ \bottomrule
\end{tabular}%
}
\caption{SDI results with interventions at conclusion-last token position (Con-$\ell$) and final token position (Last). Lower numbers indicate that the subspace involved in property inheritance in the deductive case is sensitive to the directionality of the inference.}
\label{tab:das-directional-sensitivity}
\end{table}

\section{Additional DAS Results}
\label{sec:additional_results}

\begin{figure*}
    \centering
    \includegraphics[width=\linewidth]{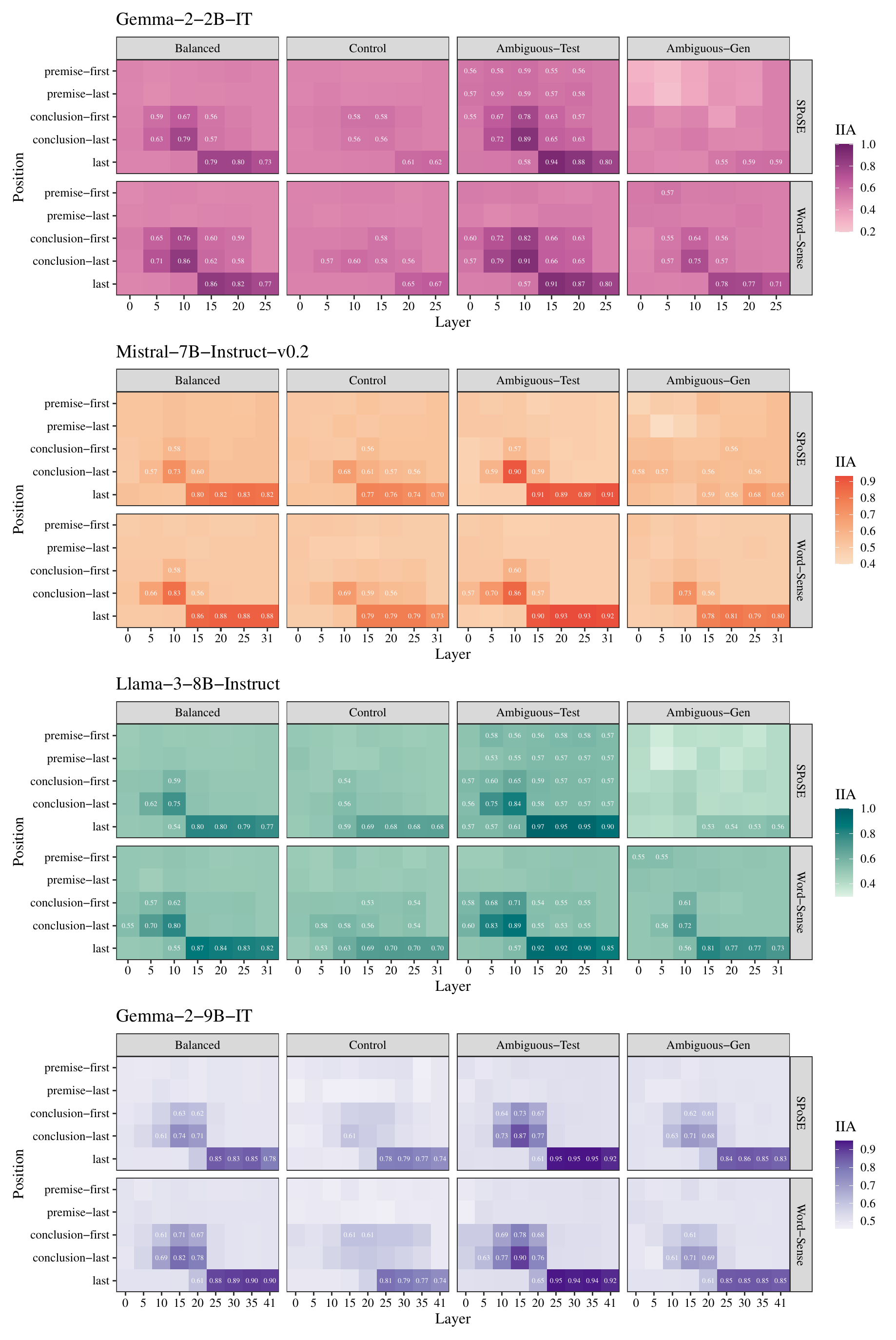}
    \caption{Interchange Intervention Accuracy (IIA) for the causal graph in Figure \ref{fig:das_diagram}, for all LMs.}
    \label{fig:das-results-all}
\end{figure*}

\begin{figure*}
    \centering
    \includegraphics[width=\linewidth]{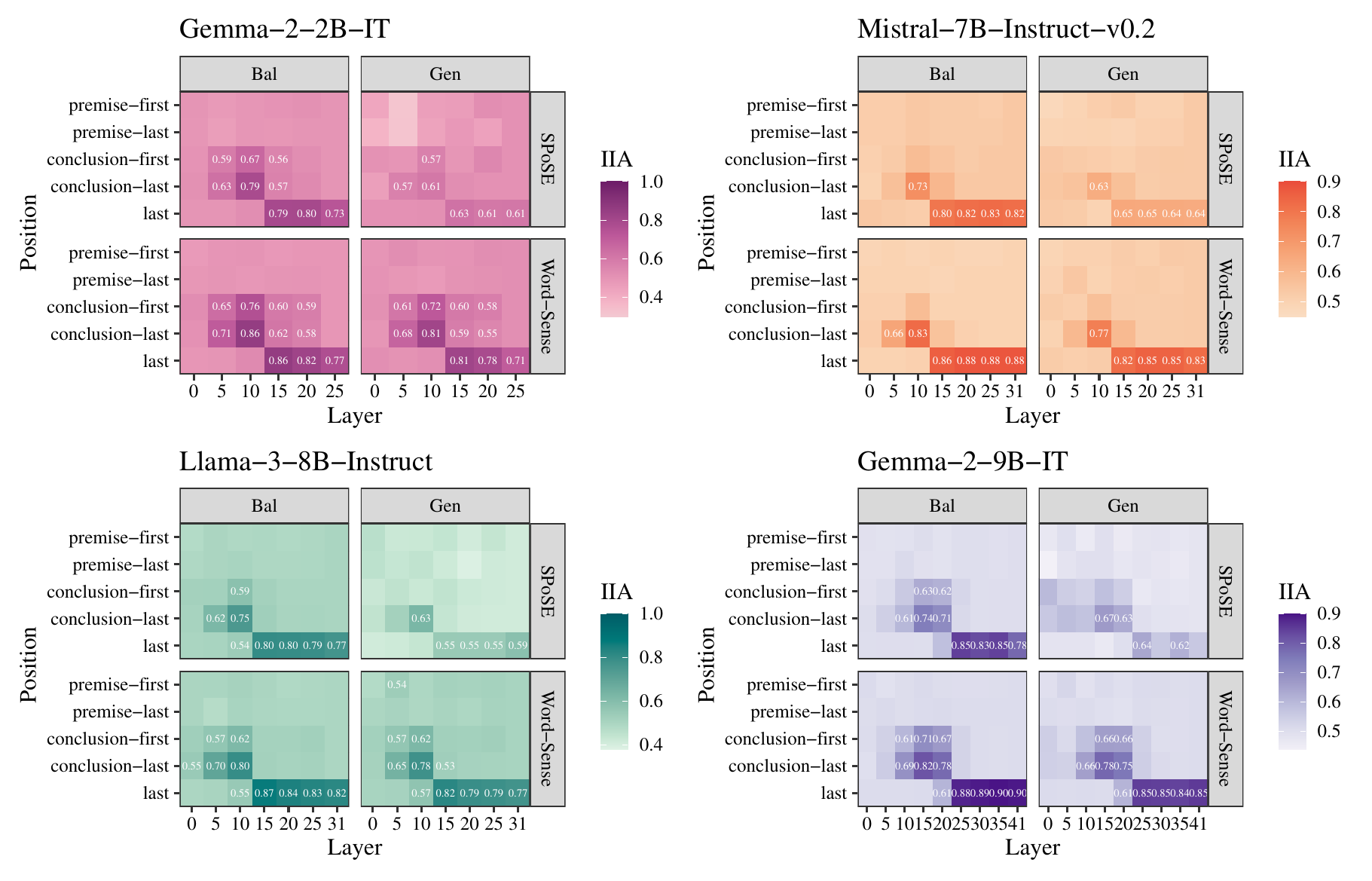}
    \caption{Interchange Intervention Accuracies (IIA) obtained on the Balanced and Generalization sets in the Unambiguous experiments, for all LMs.}
    \label{fig:unambig-results}
\end{figure*}

Additional results for the DAS experiments for Gemma-2-2B-IT and Llama-3-8B-Instruct are given in \Cref{fig:das-results-all}. This figure also includes the full set of results from using the sense-based similarity measure to derive our negative samples. 
The full set of results from the \textbf{Unambiguous} DAS experiments are shown in \Cref{fig:unambig-results}.

\section{Prompts}
\label{sec:prompts}

The formats for the prompts used in our experiments are given below.

\vspace{.5em}
\begin{displayquote}
    \small
    \underline{Prompt 1:} \texttt{Answer the question. Given that \textbf{A} is/are daxable, is it true that \textbf{B} is/are daxable? Answer with Yes/No.\textbackslash n}
\end{displayquote}
\vspace{.5em}

\vspace{.5em}
\begin{displayquote}
    \small
    \underline{Prompt 2:} \texttt{Answer the question. Given that \textbf{A} is/are daxable, is it true that \textbf{B} is/are daxable? Answer with Yes/No. The answer is:}
\end{displayquote}
\vspace{.5em}

\vspace{.5em}
\begin{displayquote}
    \small
    \underline{Prompt 3:} \texttt{Answer the question. Given that \textbf{A} is/are daxable, is it true that \textbf{B} is/are daxable?<0x0A>Answer with Yes/No.<0x0A>}
\end{displayquote}
\vspace{.5em}

\vspace{.5em}
\begin{displayquote}
    \small
    \underline{Prompt 4:} \texttt{Given that \textbf{A} is/are daxable, is it true that \textbf{B} is/are daxable? Answer with Yes/No:}
\end{displayquote}
\vspace{.5em}

Table~\ref{tab:prompts-for-lms} indicates which the prompts used for each language model and whether the chat template\footnote{\url{https://huggingface.co/docs/transformers/main/en/chat_templating}} was used. These selections were made on the basis of the models' behavioral taxonomic sensitivities.

\begin{table}[h]
\centering
\resizebox{0.4\textwidth}{!}{%
\begin{tabular}{@{}lcc@{}}
\toprule
\textbf{Model} & \textbf{Prompt} & \textbf{Chat Template} \\ \midrule
Gemma-2-2B-IT & Prompt 1 & No \\
Mistral-7B-Instruct-v0.2 & Prompt 3 & No\\
Llama-3-8B-Instruct & Prompt 2 & No \\
Gemma-2-9B-IT & Prompt 2 & Yes \\ \bottomrule
\end{tabular}%
}
\caption{Prompt formats used in our experiments.}
\label{tab:prompts-for-lms}
\end{table}

\section{DAS Implementation Details}
\label{sec:das-details}

Boundless DAS interventions were trained for 2 epochs, with a batch size of 16 on NVIDIA RTX A6000 and NVIDIA A40 GPUs. We used the Adam optimizer with a learning rate of \texttt{1e-3}, including gradient accumulation and a temperature schedule.\footnote{As used in \url{https://github.com/stanfordnlp/pyvene/blob/main/tutorials/advanced_tutorials/Boundless_DAS.ipynb}} All DAS experiments were performed using the \texttt{pyvene} library, version 0.1.1. 
We loaded all models using the default \texttt{huggingface} configuration, except for Gemma-2-9B-IT, which we loaded with torch\_dtype \texttt{torch.bfloat16} due to memory constraints.
For the \textbf{Balanced}, \textbf{Unambiguous} and \textbf{Control} experiments, we trained on 3,000 stimuli and evaluated on 1,018 stimuli. For the \textbf{Unambigious} setting we trained on 1,527 stimuli for the Word-Sense dataset and 2,017 stimuli for the SPoSE dataset. The \textbf{Ambiguous-Test} set contained 536 stimuli for Word-Sense and 671 for SPoSE, while the \textbf{Ambiguous-Gen} test set containd 482 stimuli for Word-Sense and 347 for SPoSE.

\section{Premise and Conclusion Concepts}
\label{sec:concept-examples}

The list of all premise categories used in our experiments is shown in Table~\ref{tab:premise-categories}. Tables \ref{tab:spose-gemma-examples} and \ref{tab:sense-gemma-examples} show examples from different slices of our data using both SPoSE and sense-based embeddings.

\begin{table}[h]
\centering
\resizebox{\columnwidth}{!}{%
\begin{tabular}{@{}lccc@{}}
\toprule
\textbf{Slice} & \textbf{Pair} & \textbf{Similarity} & $P_{\texttt{rel}}(\text{Yes})$ \\ \midrule
\begin{tabular}[c]{@{}l@{}}Taxonomically related,\\ High similarity\end{tabular} & \textit{garden tool $\rightarrow$ shovel} & 0.83 & 0.99 \\ \midrule
\begin{tabular}[c]{@{}l@{}}Taxonomically related,\\ Low similarity\end{tabular} & \textit{garden tool $\rightarrow$ hose} & 0.72 & 0.76 \\ \midrule
\begin{tabular}[c]{@{}l@{}}Non taxonomically related,\\ High similarity\end{tabular} & \textit{garden tool $\rightarrow$ spear} & 0.81 & 0.55 \\ \midrule
\begin{tabular}[c]{@{}l@{}}Non taxonomically related, \\ Low similarity\end{tabular} & \textit{garden tool $\rightarrow$ ham} & 0.13 & 0.01 \\ \bottomrule
\end{tabular}%
}
\caption{Examples of the four slices in our data, where similarity is calculated using \spose{} \citep{zheng2019revealing, THINGSdata}, and the model responses are from \textbf{Gemma-2-9B-IT}.}
\label{tab:spose-gemma-examples}
\end{table}

\begin{table}[h]
\centering
\resizebox{\columnwidth}{!}{%
\begin{tabular}{@{}lccc@{}}
\toprule
\textbf{Slice} & \textbf{Pair} & \textbf{Similarity} & $P_{\texttt{rel}}(\text{Yes})$ \\ \midrule
\begin{tabular}[c]{@{}l@{}}Taxonomically related,\\ High similarity\end{tabular} & \textit{sea animal $\rightarrow$ whale} & 0.79 & 0.99 \\ \midrule
\begin{tabular}[c]{@{}l@{}}Taxonomically related,\\ Low similarity\end{tabular} & \textit{sea animal $\rightarrow$ coral} & 0.70 & 0.74 \\ \midrule
\begin{tabular}[c]{@{}l@{}}Non taxonomically related,\\ High similarity\end{tabular} & \textit{sea animal $\rightarrow$ aquarium} & 0.75 & 0.42 \\ \midrule
\begin{tabular}[c]{@{}l@{}}Non taxonomically related, \\ Low similarity\end{tabular} & \textit{sea animal $\rightarrow$ rack} & 0.45 & 0.05 \\ \bottomrule
\end{tabular}%
}
\caption{Examples of the four slices in our data, where similarity is calculated using the sense-based embeddings \citep{loureiro2022lmms}, and the model responses are from \textbf{Gemma-2-9B-IT}.}
\label{tab:sense-gemma-examples}
\end{table}

\begin{table}[]
\centering
\begin{tabular}{@{}ll@{}}
\toprule
\textbf{premise category} & \textbf{$N$} \\ \midrule
food & 291 \\
animal & 177 \\
tool & 142 \\
clothing & 107 \\
container & 105 \\
mammal & 88 \\
electronic device & 74 \\
vehicle & 70 \\
weapon & 48 \\
plant & 47 \\
home decor & 45 \\
vegetable & 42 \\
accessory & 37 \\
dessert & 36 \\
furniture & 36 \\
breakfast & 35 \\
fruit & 34 \\
musical instrument & 33 \\
toy & 33 \\
fastener & 31 \\
toiletry & 31 \\
auto part & 30 \\
sea animal & 30 \\
bird & 28 \\
kitchen tool & 27 \\
medical equipment & 26 \\
school supply & 26 \\
office supply & 24 \\
seafood & 24 \\
kitchen equipment & 20 \\
drink & 19 \\
game & 19 \\
headwear & 19 \\
water vehicle & 19 \\
women's clothing & 19 \\
livestock & 18 \\
garden tool & 17 \\
insect & 17 \\
outerwear & 16 \\
protective clothing & 16 \\
candy & 15 \\
condiment & 15 \\
footwear & 15 \\
jewelry & 15 \\ \bottomrule
\end{tabular}%
\caption{List of premise categories and their sizes (in terms of number of members), sorted by size. Many of these categories overlap---e.g., mammals are also animals, etc.}
\label{tab:premise-categories}
\end{table}

\end{document}